# Deep Learning Based Segmentation Free License Plate Recognition Using Roadway Surveillance Camera Images

Alperen Elihos
Havelsan Inc.
aelihos@
havelsan.com.tr

Burak Balci
Havelsan Inc.
bbalci@
havelsan.com.tr

Bensu Alkan
Havelsan Inc.
balkan@
havelsan.com.tr

Yusuf Artan
Havelsan Inc.
yartan@
havelsan.com.tr

*Abstract* - Smart automated traffic enforcement solutions have been gaining popularity in recent years. These solutions are ubiquitously used for seat-belt violation detection, red-light violation detection and speed violation detection purposes. Highly accurate license plate recognition is an indispensable part of these systems. However, general license plate recognition systems require high resolution images for high performance. In this study, we propose a novel license plate recognition method for general roadway surveillance cameras. Proposed segmentation free license plate recognition algorithm utilizes deep learning based object detection techniques in the character detection and recognition process. Proposed method has been tested on 2000 images captured on a roadway.

## INTRODUCTION

Recently proposed computer vision and deep learning techniques have aroused a major interest in the intelligent transportation community. In recent years, automated solutions have been proposed towards traffic enforcement tasks such as seat belt violation detection, driver cell phone usage violation detection [1].

These automated enforcement solutions rely on robust license plate recognition algorithms to improve operational efficiency. Existing license plate recognition systems typically utilizes character segmentation techniques in the license plate recognition process. For a robust character segmentation, high resolution cameras are typically employed in the automated license plate recognition applications. In this study, we examine license plate recognition performance for a roadway surveillance camera using a character segmentation free algorithm.

Early works on license plate recognition typically utilizes a vertical and horizontal projection operation to extract characters within the localized license plate region [2]. This approach typically is prone to error in the presence of shadow and lighting changes. In order to overcome errors caused by the vertical & horizontal projection operations, several studies have proposed character detection based approach to localize characters. These algorithms are known as segmentation free approach to license plate recognition. In one study, Bulan *et al.* [3] proposed a Deformable Part Model (DPM) based character detector to localize characters within the license plate region. In another study, [4] proposed a convolutional neural network model to recognize Chinese characters in the license plate. Similarly, [5] proposed a Convolutional Neural Network (CNN) model to recognize Latin characters in the license plates. Li *et al.* [6] proposed a CNN based character recognition and Long Short Term Memory (LSTM) based spatial information encoding framework towards license plate recognition task. In a recent study, [7] presented the most similar approach to ours, in which a deep learning based character model is constructed using You Only Look Once (YOLOv2) object detector to localize and to recognize characters in the license plate. Rather than using YOLOv2 model to construct character models, we have utilized Single Shot Multi-Box Detector (SSD) [8] to localize and to recognize license plate characters. Proposed approach works end-to end in terms of license plate region detection and license plate recognition.

Figure 1 illustrates the overview of our proposed approach to license plate recognition. For a given roadway surveillance camera image, license plate region is localized using SSD based license plate detector. Then we run SSD character detector to detect and recognize characters on the license plate. Proposed solution produces highly robust results to lighting variations and runs relatively fast at 8 fps.

Unlike the earlier work on deep learning based character recognition, we construct models using artificially generated dataset [5]. In the next section, we present the details of our methodology. Then, we report our experiments and results using real world images. Final section presents our conclusion.

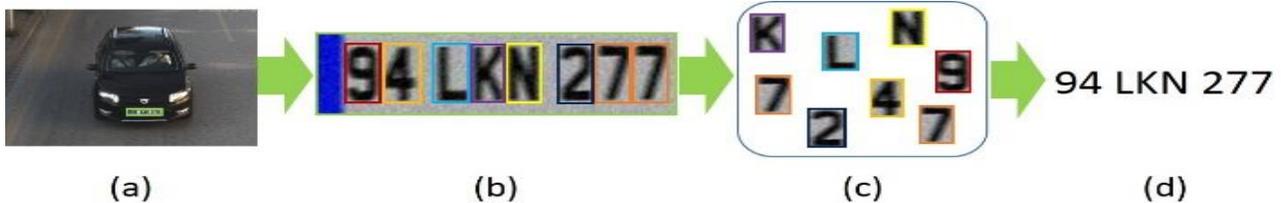

FIGURE 1. OVERVIEW OF THE PROPOSED METHOD. (A) INPUT IMAGE, (B) LICENSE PLATE LOCALIZATION, (C) CHARACTER DETECTION, (D) LICENSE PLATE RECOGNITION.

## METHODOLOGY

In this section, the details of the proposed solution for license plate recognition are described. The Proposed solution consist of three parts; License plate detection, character detection and license plate recognition. Figure 1 illustrates these parts.

In this study, we utilize SSD, which is a popular deep learning based object detection technique, for our object detection purposes [8]. In terms of object detection tasks, SSD model is shown to perform better than alternatives (Faster R-CNN [9] and YOLO [10]) in terms of speed and accuracy [11]. SSD model searches objects in feature maps from various layers that makes it able to detect various sized objects. Using the SSD object detector, we perform license plate region and character detection as explained next.

### I. License Plate Localization

The first step in the license plate recognition task is the localization of license plate on the incoming vehicle within the captured image. License plate region is the main region of interest in license recognition task. Since the remaining part of the image is irrelevant for our task, this part is ignored. For license plate detection operation, SSD model [8] is utilized. The model is constructed using a license place region annotated training dataset.

### II. Character Detection

On the detected license plate region, license plate characters are localized with an object detector. In this stage, we compare SSD object detection method with DPM [12], which is an effective model for character detection on license plates as shown in [3]. The trained models can detect 33 different characters; 23 English letters (excluding 'Q', 'W' and 'X') and 10 numbers from '0' to '9'. The details of these methods are as follows:

SSD Model: In this approach, SSD object detector [8] is utilized to detect the characters within the input image. For this operation a character detection SSD model is trained using a character regions annotated training dataset.

DPM Model: We utilized a deformable part based character detection model in which each part is a node on the tree (we used 3 nodes in the tree) and mixture model captures the structure of the 33 different characters.

### III. License Plate Recognition

After the character detection task, license plate is recognized. For an accurate recognition, some rules are obtained. The first rule is, the detected character is ordered with respect to their center pixel points. The second rule is, if the detected character region is overlapped with another detected character region and this overlapping ratio is greater than 70 %, the one with the highest detection score is used as the detected character. The final rule is, the first two and the last two character of the plate should be number. Thus, any letter detected on this range is ignored.

## EXPERIMENTS

### I. Image Acquisition

In this study, a 2MP (1920x1080) RGB and a 2MP (1920x1080) NIR camera pair with the same field of view (FOV) are placed on an overhead gantry approximately 4.5 m above the ground level.

### II. Dataset

In the training stage for license plate detection, we utilized 3000 annotated real life images (1500 NIR + 1500 RGB) to learn parameters of the classification model and 750 images were used for validation. For testing purpose, 2000 real life images were used. Figure 2 shows several sample images from our plate detection dataset.

For the training of character detection model, we utilized 72000 annotated synthetic plate images (36000 NIR + 36000 RGB) to learn parameters of the classification model and 18000 images were used for validation. For testing purpose, the same 2000 real life images, which are used for testing the plate detection model, were used. However, in this scenario, only the plate region of these images were used. Figure 3 shows several sample images from our character detection dataset.

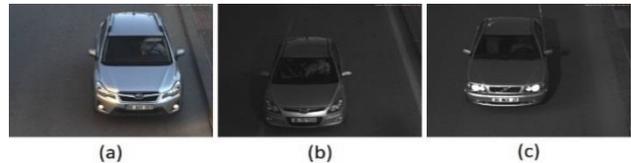

FIGURE 2. VISUAL ILLUSTRATION OF PLATE DETECTION DATASET. (A) DAY-TIME RGB IMAGE (B) DAY-TIME NIR IMAGE (C) NIGHT-TIME NIR IMAGE.

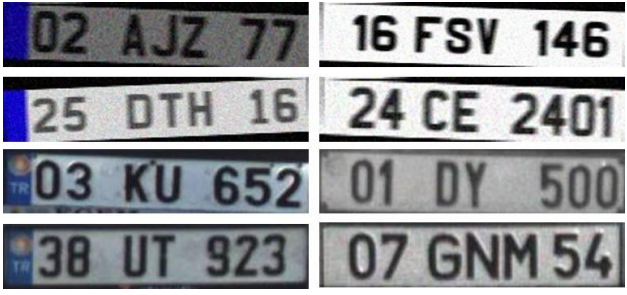

FIGURE 3. VISUAL ILLUSTRATION OF CHARACTER DETECTION DATASET. THE FIRST TWO AND THE LAST TWO ROWS REPRESENT SYNTHETIC AND REAL LIFE PLATE IMAGES RESPECTIVELY. THE FIRST AND THE SECOND COLUMN SHOWS RGB AND NIR IMAGES RESPECTIVELY.

*III. Training*

In this study, either a single channel NIR or a three channel RGB image are utilized in the decision making process. Instead of creating different models for two types of image source, we convert single channel NIR images to 3 channel NIR images by cloning them channel-wise and generate a single model using an NIR or an RGB image. Below, we outline procedures and hyper parameter selections for SSD models in license plate detection and character detection stage and for DPM model in the character detection stage.

SSD Model: During the training process, we utilized transfer learning approach to make the training process more efficient. We utilize a base SSD model presented in [8]. Using this base model, we fine-tuned it with our specific datasets. Fine tuning operation is performed by freezing the weights of the first three convolutional blocks of the model. The rationale behind this strategy is based on two facts. First three convolutional blocks trained with a large dataset (ImageNet-1k dataset [13]) behave as a feature extractor. Thus, there is no need to update these weights with our relatively small dataset. Secondly, since the first feature map to be analyzed to detect objects fall into fourth convolutional block, it is logical to update weights starting from there. In our fine tuning operations, we set the batch size as 16. As learning hyper parameters, Adam optimizer with a relatively small learning rate 0.0003 is utilized. Also we applied learning rate decay strategy shown in Eq. 1 where $\lambda$ is the learning rate, i is the epoch number.

$$\lambda_{i+1} = \lambda_i * 0.9 \quad (1)$$

DPM Model: Proposed tree model $T = (V, E)$ is a pictorial structure where V is the set of parts, and E is the set of edges between parts. [14] defined a score for a particular configuration of parts $L = \{l_i\}$, for a given image $I$ as shown in Eq. (2), where $\varphi$ is the histogram of gradients features (for the landmark points) extracted at pixel

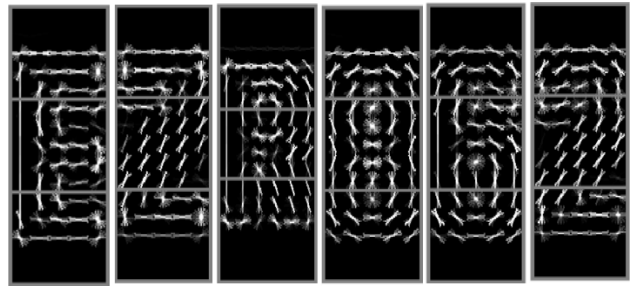

FIGURE 4. DPM MODELS FOR LETTERS 'E', 'Z', 'R' AND NUMBERS '8', '6', '2'.

$$S(I, L) = \sum_{i \in V} w_i \cdot \varphi(I, l_i) +$$

$$\sum_{i,j \in E} a_{ij} dx^2 + b_{ij} dy^2 + c_{ij} dx + d_{ij} dy \quad (2)$$

location $l_i = (x_i, y_i)$. First term sums the appearance evidence for placing the $i^{th}$ template, $w_i$ at location $l_i$. Second term score the spatial arrangement of the set of parts $L$, where $dx$ $(dy)$ term represents the spatial deformation in x (y) axis between parts $i$ and $j$. This model can be viewed as a linear classifier [15] with unknown parameters $w$ and deformation parameters $\{a, b, c, d\}$ learned during training using latent SVM as shown in [14]. Figure 4 illustrates the DPM models obtained for several characters. For a given test image $I$, we maximize Eq. (3) using dynamic programing to find the best configuration of parts.

$$S^*(I) = \max_m [\max_L S(I, L)] \quad (3)$$

*IV. Results*

In terms of the detection evaluation criterion for plate detection model, if the overlap between ground truth and detector output is greater than 80 %, the detection is assumed as correct. By using this criterion, 1912 images out of 2000 test images were detected correctly. This shows that, license plate detection model achieved 95.6 % accuracy rate.

Once the license plate region is detected, the character detection methods and license plates recognition rules are applied on the outputs of plate detection method. Table 1 represents the accuracy results of SSD and DPM. From Table 1, it is clear that SSD gives better performance than DPM. Thanks to spatial and spectral learning mechanism of SSD, it recognized the pattern of characters better than DPM. Therefore it gives better performance than DPM. Since the trained DPM model tries to capture the structure of characters with 3 nodes, it has some trouble on differentiating some character pairs such as '0 and D', 'Y and V', 'S and 8' and '3 and 9'. Because of that confusion, DPM model did not recognize the majority of the plates correctly.

TABLE 1. ACCURACY RATE OF THE COMPARED METHODS.

| Methods | SSD | DPM |
|---|---|---|
| Accuracy | **0.733** | 0.470 |

TABLE 2. A VISUAL ILLUSTRATION OF SSD OUTPUT FOR SEVERAL REAL PLATE IMAGES. GREEN RESULTS DENOTES THE CORRECT DETECTIONS, RED RESULTS DENOTES THE INCORRECT DETECTIONS.

| PLATES | SSD RESULTS |
|---|---|
| 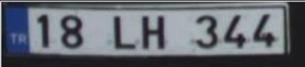 18 LH 344 | 18 LH 344 |
| 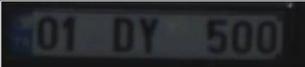 01 DY 500 | 01 DY 500 |
| 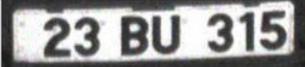 23 BU 315 | 23 BU 315 |
| 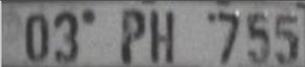 03 PH 755 | 03 PII 756 |
| 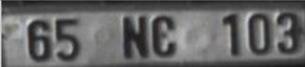 65 NE 103 | 65 NE 103 |

Additionally, computation time of license plate recognition by using SSD is analyzed on a computer with 16 GB RAM, Intel Core i7 processor and an Nvidia GeForce GTX 780 Ti GPU card. The timings are averaged over 2000 images. Considering the license plate and character detection, license plate recognition lasts 154 milliseconds on an image.

Finally, visual illustration of the results for license plate recognition by using SSD is presented in Table 2. Note that if the characters of plate are visible and plate region is detected correctly, then we are able to make a correct decision as shown in the first three rows. In $4^{th}$ and $5^{th}$ row, we see a wrong decision due to a weak character signal and a screw, which is at the center of the letter 'C', respectively.

## CONCLUSION

In this study, we proposed license plate recognition method using general roadway surveillance camera images. Proposed segmentation free method utilizes state-of-the-art deep learning based object detection technique. In order to compare our deep learning based SSD model, we utilized DPM model. Proposed SSD model typically achieve an overall accuracy around 73.3 % on a test set consisting of 2000 real life images. In the future, we will look into using the combination of different deep learning based object classification techniques and well-known classification techniques.